# Cellular Automata Segmentation of the Boundary between the Compacta of Vertebral Bodies and Surrounding Structures

Jan Egger[a,b,c] and Christopher Nimsky[a]

[a] University of Marburg, Department of Neurosurgery, Baldingerstrasse, 35033 Marburg, Germany
[b] TU Graz, Institute for Computer Graphics and Vision, Inffeldgasse 16, 8010 Graz, Austria
[c] BioTechMed, Krenngasse 37/1, 8010 Graz, Austria

## ABSTRACT

Due to the aging population, spinal diseases get more and more common nowadays; e.g., lifetime risk of osteoporotic fracture is 40% for white women and 13% for white men in the United States. Thus the numbers of surgical spinal procedures are also increasing with the aging population and precise diagnosis plays a vital role in reducing complication and recurrence of symptoms. Spinal imaging of vertebral column is a tedious process subjected to interpretation errors. In this contribution, we aim to reduce time and error for vertebral interpretation by applying and studying the GrowCut-algorithm for boundary segmentation between vertebral body compacta and surrounding structures. GrowCut is a competitive region growing algorithm using cellular automata. For our study, vertebral T2-weighted Magnetic Resonance Imaging (MRI) scans were first manually outlined by neurosurgeons. Then, the vertebral bodies were segmented in the medical images by a GrowCut-trained physician using the semi-automated GrowCut-algorithm. Afterwards, results of both segmentation processes were compared using the Dice Similarity Coefficient (DSC) and the Hausdorff Distance (HD) which yielded to a DSC of 82.99±5.03% and a HD of 18.91±7.2 voxel, respectively. In addition, the times have been measured during the manual and the GrowCut segmentations, showing that a GrowCut-segmentation – with an average time of less than six minutes (5.77±0.73) – is significantly shorter than a pure manual outlining.

## 1. DESCRIPTION OF PURPOSE

The back is the posterior part of the body's trunk from the neck to the pelvis, and is an intricate structure consisting of bones, muscles, and other tissues[1]. A major public health concern is osteoporosis – a decrease in bone mass and density –, and the urgency for accurate diagnosis of vertebral fractures has recently be shown in clinical and epidemiologic trials on osteoporosis[2]. The lifetime risk of osteoporotic fracture is 40% in white women and 13% in white men in the United States, and overall 30 million American women and 14 million men are affected by osteopenia or osteoporosis[3,4]. Vertebral fractures often affect activities of daily living, like walking, taking stairs, getting up from a chair, bathing, dressing, and cooking[5-7]. With increasing incidence of vertebral bone disease and resulting limitation of mobility and quality of life of older patients has led to increased number of spinal surgical procedure within this age group; and when a decision for an adequate procedure is made neuro-imaging plays the main role for estimating the dimension of surgery[8]. Thereby, an objective and accurate analysis of vertebral deformations is of significant importance for clinical diagnostics and therapy of pathological conditions affecting the spine[9]. In brief, the aim of a computer-assisted diagnosis (CAD) system is to (1) facilitate characterization and quantification of abnormalities, and (2) minimize interpretation errors caused by tedious tasks of image screening and radiologic diagnosis[10]. However, amongst others, the variations in soft tissue contrast makes the segmentation of vertebral bodies in MR images a challenging task[11]. In this contribution, our aims are to illustrate in general the GrowCut method for vertebral body segmentation, and to perform semi-automated segmentation of vertebrae images derived from T2-weighted Magnetic Resonance Imaging (MRI) acquisitions and demonstrate that it can speed-up a pure manually slice-by-slice analysis. Thereby, we are segmenting the boundary between the compacta of vertebral bodies and the surrounding structures.

This contribution is organized as follows: Section 2 presents details of the methods, Section 3 discusses experimental results and Section 4 concludes the paper and gives an outlook on future work.

---
E-mail: jan.egger@icg.tugraz.at

## 2. METHODS

Vertebral bodies from different subjects have been segmented in diagnostic T2-weighted magnetic resonance imaging scans of the vertebral column for this study. The datasets have been acquired on a 1.5 Tesla MRI scanner from Siemens (MAGNETOM Sonata) with a slice thickness of 4 mm. Afterwards, for a consistent comparison for the evaluation, the datasets have been reformatted to isotropic resolutions (twice to 0.63x0.63x0.63 mm$^3$ and once to 0.73x0.73x0.73 mm$^3$) to sizes of 512x512x113, 512x512x113 and 512x512x70. The images have previously been used in[12-17] and the freely available datasets can be found here:

http://www.cg.informatik.uni-siegen.de/de/spine-segmentation-and-analysis

The manual segmentations of each vertebral body were performed on a slice-by-slice basis by neurosurgeons at the University Hospital of Marburg (UKGM) in Germany. The neurosurgeons had several years of experience in the treatment of vertebral diseases and the three-dimensional manual slice-by-slice segmentations were performed in the sagittal direction with corrections in axial and coronal directions. However, if the vertebral border was very similar between consecutive slices, the contouring software allowed the user to skip manual segmentation in each slice, and instead interpolated the boundaries in these areas. The software used for the manual contouring was created within MeVisLab[18-23]. However, the software provided only simple contouring capabilities without any algorithmic support to avoid falsifying the results. Figure 1 and Figure 2 present voxelizations of manual segmentations in 2D and 3D. The upper images of Figure 1 show manual contours (green) on the left side and the corresponding voxelized vertebra mask on the right side (gray). The lower image of Figure 1 shows on the left side a single manual contour (green) in a sagittal slice. The lower right image of Figure 1 shows the corresponding voxelized vertebra mask (yellow) at the same position on the sagittal slice. Figure 2 presents manual segmentations of seven vertebral bodies (yellow) in 3D with a superimposed sagittal MRI slice. Amongst others, these vertebral bodies have been used for evaluation in this study.

Similar to [24-26] for glioblastoma multiforme (GBM)[27-29], Pituitary Adenoma[30,31] and lung cancer, the software used during this study for the semi-automatic segmentation work was Slicer (www.slicer.org)[32]. After testing various segmentation tools available within the Slicer platform, like the Robust Statistics Segmenter (RSS), we concluded that the use of GrowCut followed by additional morphological operations (like dilation, erosion and island removal) provides the most efficient segmentation method for our datasets. Thus, the following step-by-step workflow to perform vertebral body segmentation has been used (Note: The physicians were also trained to segment the vertebral bodies according to this workflow):

- loading the patient dataset into the Slicer Platform;
- initializing the foreground and background for GrowCut, by marking an area inside and around the identified vertebral bodies;
- running the automatic competing region-growing in Slicer; and
- using morphological operations like dilation, erosion, and island removal for post-editing after visual inspection of the results.

Briefly, the GrowCut segmentation method is a competitive region-growing algorithm using cellular automata that uses an iterative labeling procedure. Thus, a GrowCut-based segmentation achieves reliable and reasonably fast segmentations of moderately difficult objects in 2D and 3D. The algorithm uses a set of input pixels for the foreground and the background and then tries to label all the pixels in the image iteratively. The algorithm stops when all the pixels in the Region of Interest (ROI) have been labeled, and no pixel can change its label any more. However, the GrowCut implementation employs some techniques for speeding up the automatic segmentation process that we want to summarize here: first of all, the implementation computes the segmentation only within a small ROI, given that the user is typically only interested in segmenting out a small area of the whole image. Therefore, the ROI is computed as a convex hull of the pixels labeled by the user, plus an additional margin of about 5%. In addition, several small regions of the image can be updated simultaneously, which is achieved by executing iterations involving the image in multiple threads. Furthermore, the similarity distance between the pixels can be pre-computed once and then reused, and the current implementation keeps track of "saturated" pixels for which the weights and therefore the labels can no longer be updated.

The user interface of the Editor module, which has been used for the initialization of GrowCut for the vertebra segmentation is shown in Figure 3 on the left side. Furthermore, Figure 3 shows a L4 vertebra after the dataset is loaded with a typical user initialization for GrowCut on the axial, sagittal and coronal cross-sections on the right side. As hardware platform for this study, we used a computer with Intel Core i5-750 CPU, 4x2.66 GHz, 8 GB RAM, with

Windows XP Professional x64 Version, Version 2003, Service Pack 2. Note: An initialization of the foreground and background as shown in Figure 3, took a trained physician less than one minute.

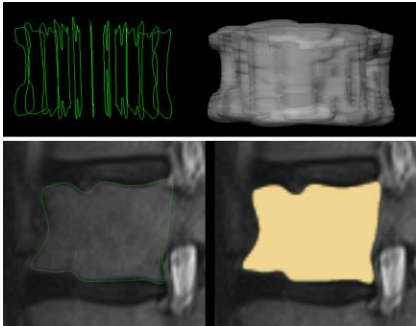
Fig. 1: 2D to 3D Voxelization.

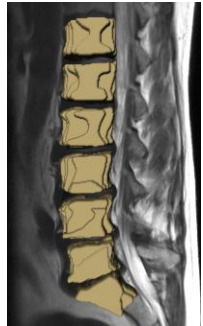
Fig. 2: 3D Voxelization.

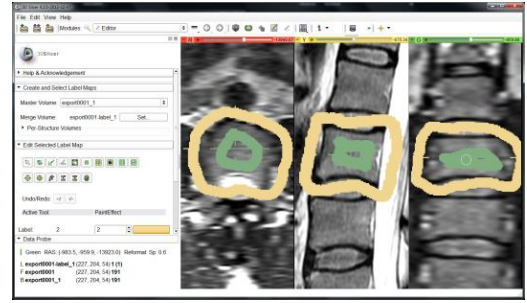
Fig. 3: Typical user initialization of GrowCut.

## 3. RESULTS

As comparison metrics for our study, the GrowCut-based vertebral body segmentation results have been evaluated against manually slice-by-slice segmentations using the Dice Similarity Coefficient (DSC) and the Hausdorff Distance. The DSC is a measure for spatial overlap of different segmentation results, which is commonly used in medical imaging studies to quantify the degree of overlap between two segmented objects A and R: $DSC = 2 \cdot V(A \cap R)/(V(A)+V(R))$. Thereby, the DSC can have a value ranging from zero to one, and is defined as two times the volume of the intersection between the two segmentations A and R, divided by the sum of the volumes of the two segmentations. A value of zero indicates no overlap and a value of one indicates a perfect agreement, and as a consequence higher values indicate a better agreement. The Hausdorff Distance is used to calculate how far away (in voxel) the two segmentations A and R are. As gold standard to calculate the DSCs and the Hausdorff Distances we had manual segmentations of vertebrae boundaries extracted by several clinical experts (neurological surgeons) with many years of experience in spine surgery. Compared with the GrowCut-based segmentation results from a trained physician we discovered an average Dice Similarity Coefficient of 82.99±5.03% and Hausdorff Distance of 18.91±7.2 voxel. Table 1 presents the summary of the results: minimum, maximum, mean $\mu$ and standard deviation $\sigma$, for all vertebral bodies.

|  | volume of vertebral body (cm$^3$) | | Hausdorff Distance (Voxel) | DSC (%) | Time (min.) |
|---|---|---|---|---|---|
|  | manual | algorithm | | | |
| min | 20.89 | 27.2 | 10.7 | 74.56 | 5 |
| max | 49.4 | 59.22 | 32.3 | 91.6 | 7 |
| $\mu \pm \sigma$ | 36.49 ± 7.15 | 44.61 ± 9.36 | 18.91 ± 7.2 | 82.99 ± 5.03 | 5.77 ± 0.73 |

Tab. 1: Summary of results: min, max, mean $\mu$ and standard deviation $\sigma$ for thirteen vertebral bodies.

For visual inspection, Figure 4 shows a GrowCut-based segmented vertebral body (green) in different views (2D and 3D): the three leftmost images show the segmentation results in 2D for an axial, sagittal and coronal plane. The next image shows a 3D view of the segmented vertebral body with an axial, sagittal and coronal plane. Finally, the image on the right presents a three-dimensional representation of a segmented L4 vertebral body with additional surface smoothing. A direct comparison of a manual (yellow) and a GrowCut-based segmentation (green) on a sagittal slice is presented in Figure 5: the left image shows the original MRI slice, the next image from the left presents the manual segmentation, the third image from the left presents the GrowCut-based segmentation result, and the right image presents both segmentations (manual and GrowCut) superimposed on the original MRI slice.

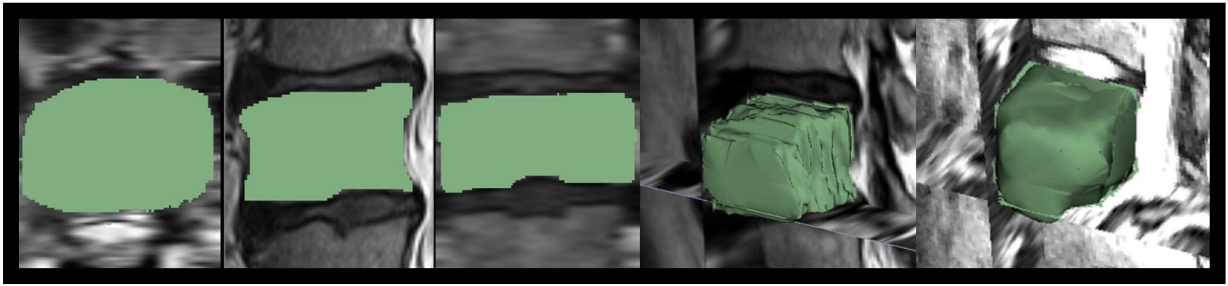

Fig. 4: GrowCut-based segmentation result for a vertebral body (green): The three leftmost images show the segmentation results in 2D for an axial, sagittal and coronal plane. The next image shows a 3D view of the segmented vertebral body with an axial, sagittal and coronal plane. Finally, the rightmost image presents a three-dimensional representation of a segmented L4 vertebral body with additional surface smoothing

.

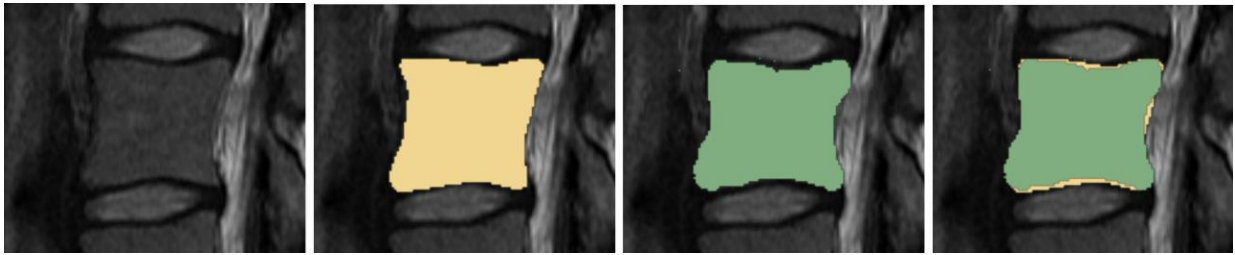

Fig. 5: Direct comparison of a manual (yellow) and the automatic segmentation (green) on a sagittal slice: The left image shows the original MRI slice, the adjacent image presents the manual segmentation, the third image from the left presents the GrowCut segmentation result and the right image presents both segmentations (manual and GrowCut) superimposed on the MRI slice.

## 4. CONCLUSIONS

In this study, we used the interactive GrowCut algorithm, based on cellular automata, for 3D segmentation of vertebral bodies (note: preliminary results have been presented at the spine congress of the DGNC in Frankfurt, Germany[33]). In summary, we found that a semi-automated segmentation using the GrowCut algorithm reduces segmentation time while at the same time achieves a similar accuracy as pure manual slice-by-slice segmentations. For evaluation of the GrowCut segmentation results, we used vertebrae images from MRI datasets, which have been manually outlined by physicians, and which took in average over ten minutes (10.75±6.65) for a single vertebra in our datasets. The physicians who generated the ground truth for the evaluation were all neurological surgeons who have several years of experience in spine surgery. However, the comparison of the automatic GrowCut segmentations with the pure manual slice-by-slice expert segmentations, resulted in an average Dice Similarity Coefficient of 82.99±5.03% and an average Hausdorff Distance of 18.91±7.2 voxel.

There are several areas of future work: The GrowCut algorithm initialization has initially been set up by the user in three slices for this study. However, instead of initializing the foreground and background on three single 2D slices, one single 3D initialization could be used by means of generating a sphere around the position of a user-defined seed point near the center of the vertebral body. Furthermore, we want to test GrowCut on longitudinal/tubular structures, like vessels[34-37], fiber tracts[38,39] or the rectum[40-42].

# ACKNOWLEDGEMENT

DDr. Jan Egger receives funding from BioTechMed-Graz ("Hardware accelerated intelligent medical imaging").